\documentclass[10pt,letterpaper]{article}
\usepackage{cogsci}

\usepackage{CJKutf8}

\newcommand{\zh}[1]{\begin{CJK}{UTF8}{gbsn}#1\end{CJK}}
\newcommand{\ko}[1]{\begin{CJK}{UTF8}{mj}#1\end{CJK}}

\usepackage{graphicx}
\usepackage{xspace}
\usepackage{balance}
\usepackage{url}
\usepackage{hyperref}
\hypersetup{
    colorlinks,
    linkcolor={red!50!black},
    citecolor={blue!50!black},
    urlcolor={blue!80!black}
}

\cogscifinalcopy 

\usepackage[
  style=apa,
  natbib=true,
  annotation=false,
]{biblatex}
\usepackage{float} 

\title{Quantifying and extending the coverage of spatial categorization data sets}

\author[1]{\mbox{Wanchun Li}}
\author[2]{\mbox{Alexandra Carstensen}}
\author[3]{\mbox{Yang Xu}}
\author[4]{\mbox{Terry Regier}}
\author[5]{\mbox{Charles Kemp}}
\affil[1]{Faculty of Engineering and Information Technology, The University of Melbourne}
\affil[2]{Psychology, Arizona State University}
\affil[3]{Department of Computer Science, Cognitive Science Program,  University of Toronto}
\affil[4]{Department of Linguistics,  University of California, Berkeley}
\affil[5]{Melbourne School of Psychological Sciences, The University of Melbourne}

\newcommand{\concept}[1]{\textsc{#1}}
\newcommand{\term}[1]{\emph{#1}}
\newcommand{\gloss}[1]{`#1'}
\newcommand{\acronym}{LCXRK\xspace}

\begin{document}

\maketitle

\begin{abstract}

Variation in spatial categorization across languages is often studied by eliciting human labels for the relations depicted in a set of scenes known as the Topological Relations Picture Series (TRPS). We demonstrate that labels generated by large language models (LLMs) align relatively well with human labels, and show how LLM-generated labels can help to decide which  scenes and languages to add to existing spatial data sets. To illustrate our approach we extend the TRPS by adding 42 new scenes, and show that this extension achieves better coverage of the space of possible scenes than two previous extensions of the TRPS. Our results provide a foundation for scaling towards spatial data sets with dozens of languages and hundreds of scenes. 

\textbf{Keywords:}
spatial semantics; large language models; cross-linguistic comparison; semantic typology; categorization
\end{abstract}

\section{Introduction}

Languages vary in how they carve the world into categories, and this
variation has been studied most extensively in the domains of
kinship~\citep[e.g.][]{jones10,kempr12,hallam2025predictive} and color~\citep[e.g.][]{kay1997color,zaslavsky2018efficient,josserand2021environment}. Progress in these domains has been enabled by the existence of standard representations, such as genealogical grids and perceptual color spaces~\citep{regierkk07}, that serve as a universal substrate for cross-linguistic comparison. The availability of these representations has allowed the development of data sets such as  Kinbank~\citep{passmorekinbank} and the World Color Survey~\citep{cookkr05} that include category
systems from more than a hundred languages.

\begin{figure*}[t]
\begin{center}
\includegraphics{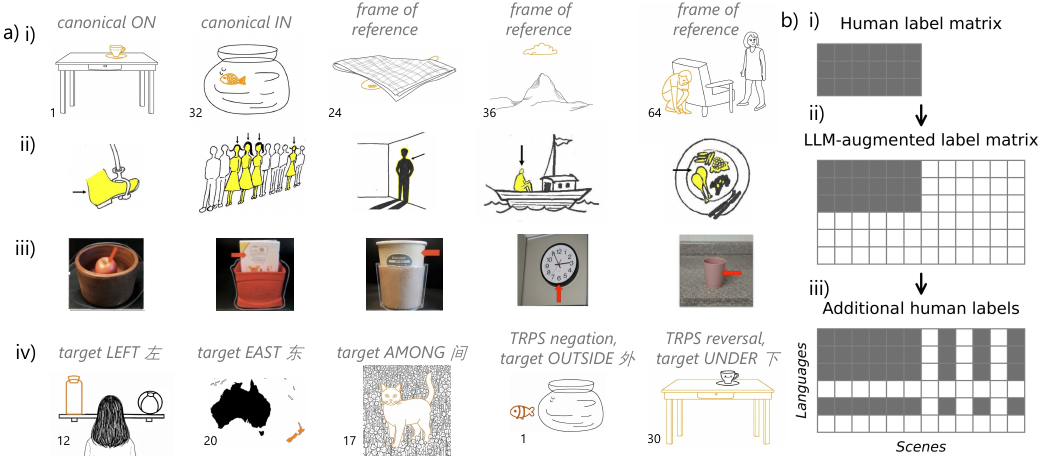}
\vspace{-0.1in}
\end{center}
\caption{(a) Spatial relations stimulus sets. Each stimulus shows the relation between a focal object (shown in gold, or marked with an arrow) and a background object. Some of the original images have been cropped for this figure. (i) The Topological Relations Picture Series \nocite{trps}(TRPS; Bowerman and Pederson, 1992) includes 71 scenes that were chosen to explore the boundaries of terms for \concept{on} and \concept{in} relations. (ii) The Zhang set~\citep{zhang13} includes 63 scenes illustrating configurations that are absent from the TRPS but relevant to the expression of \concept{on} and \concept{in} in Chinese.\protect\footnotemark[1]{} The examples here include ``girls in line,'' ``shadow on wall'' and ``food on plate.'' (iii) The LJSP set~\citep{landaujsp17} includes 44 scenes that were designed to span six subtypes of containment and five subtypes of support. The examples here include ``paper in box,'' ``cup in sleeve,'' and ``cup on counter.'' (iv) The \acronym set developed by us includes 42 images designed to illustrate spatial terms in English and Chinese that are not represented by the TRPS, and to include negations and reversals of TRPS scenes. The examples here include ``cat among flowers.'' b) Extending a cross-linguistic data set using LLMs. (i) The original human labels are organized into a dark gray matrix of languages (rows) by scenes (columns). (ii) We add candidate languages (new rows) and candidate scenes (new columns) and generate LLM labels for all new combinations of languages and scenes. (iii) A coverage measure based on LLM labels is used to prioritize languages and scenes for subsequent human labeling.}
\label{fig:stimuli}
\end{figure*}

Cross-linguistic variation in spatial categorization has also been widely studied, but this domain has been more resistant to formalization than either kinship or color. The greatest challenge is that there is no standard representation of the space of spatial relations, which makes it difficult to develop a spatial data set comparable to Kinbank or the World Color Survey.  The most widely used set of spatial stimuli is a collection of 71 images known as the Topological Relations Picture series (TRPS; Figure~\ref{fig:stimuli}a.i; \cite{trps}). As \nocite{levinsonw06} Levinson and Wilkins (2006, p 9) note, the TRPS was ``specifically designed to investigate the maximal range of scenes that may be assimilated to canonical \concept{in}- and \concept{on}-relations (and thus includes a number of scenes unlikely to be so assimilated).'' The TRPS therefore makes no attempt to cover the space of possible spatial relations, which leaves room for extended stimulus sets that achieve greater coverage of this space. 

Previous researchers have recognized the need to extend the TRPS, and \citet{zhang13} and \citet{landaujsp17} have developed new stimuli that are useful for investigating canonical \concept{in}- and \concept{on}-relations in even more detail (see Figures~\ref{fig:stimuli}a.ii and \ref{fig:stimuli}a.iii). We pursue a different goal and prioritize the notion of \text{coverage}, which we formalize as the extent to which a stimulus set is representative of the full universe of possible scenes. \citet{carstensenxkr15} suggest that greater coverage can be achieved by representing relations as lists of spatial features (e.g.\ \concept{contact}, \concept{support}, and \concept{proximity}) and sampling scenes that provide good coverage of the space of possible feature lists. This feature-based approach is appealing because it provides an explicit characterization of the space of possible relations, but relatively challenging to implement. Here we adopt the language-based approach of identifying spatial terms (e.g.\ English \term{outside} and Chinese \zh{中间}, \gloss{among}) that are not represented by the TRPS and adding scenes that can be labeled with these terms (Figure~\ref{fig:stimuli}a.iv). 


Our ultimate goal is to develop data sets that include dozens of languages and hundreds of scenes, but scaling up in this way presents a significant challenge. Previous researchers have used large language models (LLMs) to scale up psychological data sets including semantic feature norms~\citep{suresh25} and psycholinguistic databases~\citep{martinez}. We argue that LLMs can also help to identify scenes and languages that are useful to add to existing spatial relations sets (Figure~\ref{fig:stimuli}b) and illustrate our approach using LLM-generated labels for 220 scenes in 23 different languages. To justify using LLMs in this way, we first evaluate LLM labels for the original TRPS against human labels collected by \citet{carstensenkhr19} and \citet{xuk10}, and find that LLMs achieve a relatively high level of accuracy.  We do not propose that LLMs can or should replace human experimental participants, but argue that LLMs can contribute to an approach that ultimately includes human data collection. 

In what follows, we first introduce our coverage-based approach for extending existing spatial data sets and provide a formal definition of coverage. We then present and evaluate our method for eliciting spatial category labels from LLMs, and show how LLM data can help to choose which scenes and languages to add to existing data sets (Figure~\ref{fig:stimuli}b). 
We finish by discussing limitations of our approach and prospects for scaling up spatial data sets to an even greater extent. 


\section{Quantifying coverage}

\footnotetext[1]{\citet{zhang13} mentions that she developed 64 new pictures but 63 were included in the set that she kindly shared with us.}

Our general approach is summarized by Figure~\ref{fig:stimuli}b. Suppose that we already have a matrix specifying human labels for a set of scenes across a set of languages (Figure~\ref{fig:stimuli}b.i). We are considering extending the matrix to include additional scenes and languages, and use LLMs to compile labels for all of these possibilities (Figure~\ref{fig:stimuli}b.ii). We then use these LLM labels to identify the scenes and languages that increase the coverage of the original data to the greatest extent, and ultimately collect additional human labels for these scenes and languages (Figure~\ref{fig:stimuli}b.iii). Prioritizing scenes and languages based on coverage is consistent in some ways with  theory-based sampling~\citep{majid2023establishing}, or choosing a sample that varies along theoretically relevant dimensions. Prioritizing coverage, however, may also lead to samples that reveal theoretical dimensions that are exposed by the LLM labels but not yet part of theories entertained by the experimenters.

Our notion of coverage can be formalized using a similarity measure over scenes and a similarity measure over languages. Given a universe $U$ including all scenes in the LLM-augmented matrix, the coverage of
any subset $S$ of $U$ is defined as 
 the average maximum similarity between each scene in $U$ and its closest neighbour in $S$:
\begin{equation}
\text{Coverage}(S) = \frac{1}{|U|} \sum_{u \in U} \max_{s \in S} \text{sim}(s, u),
\label{eqn:coverage}
\end{equation}
where $\text{sim}(\cdot, \cdot)$ is a similarity measure over scenes. $\text{Coverage}(S)$ takes values between 0 and 1, and is high to the extent that each scene in $U$ is similar at least one scene in $S$. The same equation can be used to compute the extent to which a universe $U$ of languages is covered by a subset $S$ of languages if we use a similarity measure $\text{sim}(\cdot, \cdot)$ over languages.

Later sections explain how similarity measures over scenes and languages can be derived from the LLM-augmented matrix in Figure~\ref{fig:stimuli}b.ii. First, however, we describe our method for eliciting spatial labels from LLMs.

\section{Labeling spatial relations using LLMs}

Previous researchers have noted that AI can help scale up work on spatial categorization, and \citet{meewisfd23} compiled labels for the 71 TRPS scenes in 73 languages using a method that involves machine-translation of scene descriptions provided by English speakers. We developed a different approach in which LLMs are treated much like human participants, and directly asked to label images. 

\citet{carstensenkhr19} collected TRPS labels for at least 13 speakers of each of seven languages, and \citet{xuk10} collected labels from a single speaker for each of 21 languages. We aimed to collect LLM responses for the 23 languages represented in one or both data sets. After some initial experimentation with different LLMs and prompts, we settled
on Gemini 3 Flash for our final runs. The Gemini series is generally viewed as
handling multilingual tasks well, and as of January 2026, Gemini 3 Flash is the
top performer on MMMLU (Multilingual Massive Multitask Language Understanding),
a key multilingual benchmark. We accessed Gemini 3 Flash through its API and set the temperature parameter to zero with the goal of achieving deterministic and reproducible outputs.

We provided Gemini with a list of scenes, one
per page, and each page included a page number along with annotations
specifying the focal and background objects. For example, the page for the
top-left scene in Figure~\ref{fig:stimuli}a.i included `focal object: cup' and `background object: table.' Different stimuli sets use different visual cues for highlighting focal objects. We used a version of the TRPS that shows the focal object in gold (see Figure~\ref{fig:stimuli}a.i), and the Zhang and LJSP stimuli use yellow and red arrows respectively to indicate focal objects.

All our LLM results are based on a single run that included all 220 scenes from the four stimulus sets in Figure~\ref{fig:stimuli}a. The prompt with Chinese as a target language is as follows:
\begin{quote}
You are a native speaker of Chinese and I'd like you to respond in Chinese.
Your task is to label the spatial relationships shown in a set of images. Here
is a set of images that I'll call scenes.pdf. Each image shows a focal object
and a background object. From image 1 to image 113, the focal object is gold
and the background object is black. From image 114 to image 176, the focal
object is yellow and indicated by an arrow, and the background object is black
or blue. From image 177 to image 220, the focal object is indicated by a red
arrow. An English speaker used the following spatial terms to describe the
relationship between the focal object and the background object in each image:
1) ``on''; 2) ``in'';  ...  220) ``on''. I'd like you to label the same images in scenes.pdf. For each image in scenes.pdf, please give me the spatial term in Chinese that best describes the relationship between the focal object and the background object. For each image, please respond using a single spatial term instead of a full sentence. And please do not translate the responses I gave you in English! Instead, I'd like you to respond as a native Chinese speaker would. Your responses (one for each image in scenes.pdf) should
be organized into a numbered list. 
\end{quote}
All prompts included 220 labels provided in a reference language. Chinese was the reference when English was the target, and English was the reference for all other languages. Choosing English as the reference in most cases induces a bias, but seemed appropriate given that LLMs are likely to be more proficient in English than any other language. For the TRPS scenes, English reference labels were the modal English responses collected by \citet{carstensenkhr19}, and for the Zhang and LJSP sets, each English reference label was either \term{in} or \term{on}. For the \acronym set, the English reference labels were the target spatial terms 
that the \acronym scenes were designed to illustrate. For all four data sets the Chinese reference labels were the Chinese responses generated by Gemini 3.

\begin{figure}[t]
\begin{center}
\includegraphics{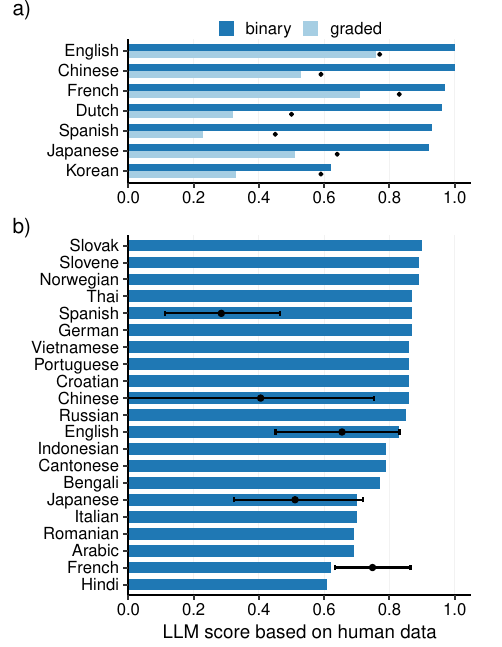}
\end{center}
\vspace{-0.2in}
\caption{Evaluation of LLM labels against data collected by (a) \protect\citet{carstensenkhr19} and (b) \protect\citet{xuk10}. The binary score for an image label is 1 if a human provided the same label, and the graded score is the proportion of humans who provided the label. In (a), black points show the maximum possible graded score for each language. In (b), black points show the average result when a single human from the \protect\citet{carstensenkhr19} data is scored with respect to a different speaker of the same language, and error bars show the 95\% interquantile range.}
\label{fig:llm_evaluation}
\vspace{-0.1in}
\end{figure}

\begin{table}[t]
\begin{center}
\caption{Coverage scores achieved by four stimulus sets relative to a universe $U$ of 220 scenes.}
    \label{tab:coverage_scores}
    \begin{tabular}{llcc}
      \hline
      Stimulus set & Size & Coverage score & 95\% CI\\
      \hline
      TRPS & 71 & 0.914 & [0.82, 0.95] \\ 
      TRPS + Zhang & 134 & 0.918 & [0.85, 0.95]\\ 
      TRPS + LJSP & 115 & 0.918 & [0.87,0.95]\\ 
      TRPS + \acronym & 113 & \textbf{0.964} & [0.96, 0.995] \\ 
      \hline
    \end{tabular}
  \end{center}
  \vspace{-0.1in}
\end{table}

Figure~\ref{fig:llm_evaluation}a compares the LLM labels with data collected by \citet{carstensenkhr19}. The binary score for a label is 1 if at least one human provided the same label, and 0 otherwise. Binary scores for six of the seven languages exceed 0.9, but the score for Korean is lower. For 15 of the 71 TRPS scenes, the LLM chooses the locative marker \ko{-에}, which does not appear in isolation in the Carstensen et al.\ data. 
Instead, Korean speakers often used verbs such as \ko{끼워져} (\gloss{fitted on}) for a ring on a finger, or \ko{걸려} (\gloss{hanging}) for a phone on the wall.

High binary scores indicate that the LLM's responses tend to match the labels of some humans, but do not reveal whether those responses are broadly typical of human labels. We therefore computed a graded score for each LLM label defined as the proportion of humans who gave the same response. Maximum graded scores for each language are shown as black points in Figure~\ref{fig:llm_evaluation}a, and the LLM averages for English, Chinese, French and Japanese are all within 0.15 of the maximum possible graded score. For Dutch and Spanish the gap between the LLM graded score and the maximum is greater. The LLM chooses \term{en} (\gloss{in, on, or at}) as the Spanish label for 41 out of 71 scenes, but Spanish speakers often give more specific responses such as \term{sobre} (\gloss{on}) for cup on table, or \term{adentro} (\gloss{inside}) for fish in bowl.

Figure~\ref{fig:llm_evaluation}b compares LLM labels with data collected by \citet{xuk10}. Only one speaker was consulted for each language, which means that the binary score for a label now indicates whether or not the LLM matched the human label for the same image. For comparison, Figure~\ref{fig:llm_evaluation}b includes black points showing the average alignment score between two speakers of the same language in the \citet{carstensenkhr19} data. The average LLM score exceeds 0.8 for 12 of the 21 languages, and French and Hindi achieve the lowest scores. The LLM scores relatively well for French in Figure~\ref{fig:llm_evaluation}a, however, suggesting that the single French speaker  in the \citet{xuk10} data set gave somewhat idiosyncratic responses, such as \term{en haut a droite de} (\gloss{on the top right of}) instead of \term{sur} (\gloss{on}) for a stamp on a letter.

For us, knowing whether the LLM produces relatively reliable labels is important, but understanding how it produces these labels is less essential. We wondered, however, whether the LLM actually needs images of the TRPS scenes to generate reliable labels. 
We therefore ran a text-based condition that was almost identical to our original image-based condition except that all images were stripped from the file \texttt{scenes.pdf} provided to the LLM, leaving only specifications of the focal and background objects in each scene. The results were similar to those reported in Figure~\ref{fig:llm_evaluation}, and average scores across languages were virtually identical: mean binary scores across the 7 languages in Figure~\ref{fig:llm_evaluation}a were 0.91 (image-based) and 0.90 (text-based), mean graded scores were 0.48 (image-based) and 0.49 (text-based), and mean binary scores across the 21  languages in Figure~\ref{fig:llm_evaluation}b were 0.80 (image-based) and 0.78 (text-based). These results therefore suggest that image analysis makes little contribution if any to the scores reported in Figure~\ref{fig:llm_evaluation}.

As of January 2026, Google Translate supports 249 languages, including all of the languages in Figure~\ref{fig:llm_evaluation}, which suggests that 249 is an upper bound on the number of languages to which our LLM-based approach can be applied. Languages outside of this set account for much of the world's linguistic diversity~\citep{joshi,grambank}, but are poorly represented in the training data available to LLMs. For high-resource languages, however, our results so far suggest that LLM-generated labels of spatial configurations are accurate enough to support research on spatial categorization

\begin{figure*}[t]
\begin{center}
\includegraphics[]{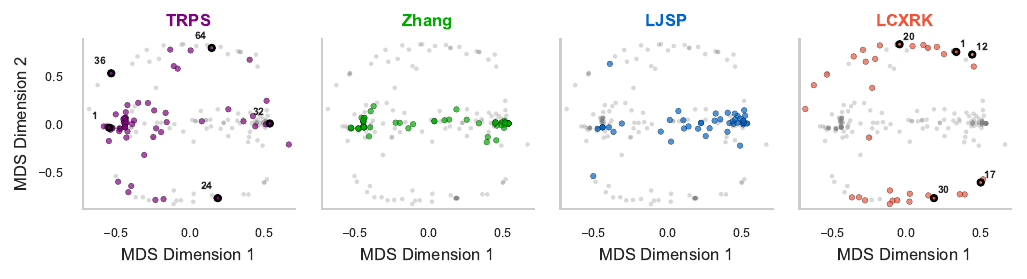}
\end{center}
\vspace{-0.2in}
\caption{MDS visualization of a space that includes scenes from all stimulus sets considered in this paper. The four panels show the coverage of this space achieved by (a) the TRPS (b) the Zhang set (c) the LJSP set and (d) our new stimulus set. Points labeled in panels (a) and (d) correspond to images shown in Figures~\ref{fig:stimuli}a.i and \ref{fig:stimuli}a.iv.}
\label{fig:scene_space}
\end{figure*}

LLMs may be especially valuable when running quick preliminary tests of hypotheses that require many languages to evaluate, or when choosing languages or stimuli to prioritize for future experiments with humans (Figure~\ref{fig:stimuli}b). The following sections illustrate the latter possibility.


\section{Adding scenes to existing data sets}

\citet{levinsonw06}  present a taxonomy that organizes spatial relations into three groups: topological relations (e.g.\ English \term{in} and \term{on}), frame-of-reference relations (e.g.\ \term{behind}, \term{left of}, \term{north of}) and relations involving motion (e.g.\ \term{from London to Sydney}). Because we work with static pictures, we do not consider motion-related relations, but we sought to add additional topological relations (e.g.\ \term{outside}, \term{off}) and frame-of-reference relations (\term{left of}) to the TRPS. 
Despite its name, the TRPS is not restricted to topological relations. It includes frame-of-reference relations such as ``spoon under cloth'' 
and ``boy behind chair,'' which help to characterize how languages treat the boundary between topological and frame-of-reference relations. Adding additional frame-of-reference relations to the TRPS therefore builds on a theme that is already present in the original data set.

We used two distinct strategies to add 42 scenes to the TRPS. 27 scenes were added by identifying spatial terms in some language that are not already represented by the TRPS, and designing new scenes that illustrate these relations. The authors include native speakers of Chinese and English, and we therefore focused on these two languages. We consulted lists of Chinese and English spatial terms provided by ~\citet{tyler2003semantics}, \citet{ linguistic_terms_2011} and \citet{landauj93}, and identified 11 Chinese terms (\zh{外}, \zh{间}, \zh{中}, \zh{东}, \zh{西}, \zh{南}, \zh{北}, \zh{左}, \zh{右}, \zh{前}, \zh{后}) and 15 English terms (\term{across, after, along, among, at, between, down, east, left, north, off, outside, right, south, west}) that seemed like good candidates for illustration. We chose focal and background objects for all of these cases, then prompted Gemini to generate illustrations like the examples in Figure~\ref{fig:stimuli}a.iv. 

The remaining 15 scenes were created by using two methods to adjust existing TRPS stimuli. The first method negates the original TRPS relation (e.g.\ so that a fish is outside a fishbowl instead of in the fishbowl). The second method keeps the original TRPS scene but exchanges the focal and background objects (e.g.\ so that the table is under the cup instead of the cup being on the table). We did not exhaustively apply these transformations to all scenes, but included 15 cases that seemed sensible to us.

Thousands if not millions of scenes could be added to the TRPS, and a coverage measure is useful for identifying which scenes should be prioritized when collecting human labels. We used the coverage measure in Equation 1 to compare the coverage of three extensions of the TRPS. The first is the \acronym set, which consists of our 42 new scenes, and the second and third are the Zhang and LJSP sets shown in Figures~\ref{fig:stimuli}a.ii and \ref{fig:stimuli}a.iii.\footnote[2]{Unfortunately the work of \citet{carstensenxkr15} could not be included as a fourth set because those authors no longer have a copy of their stimuli} Both of these projects focused on \concept{on} and \concept{in} relations, and achieving broad coverage of topological and frame-of-reference relations was a goal of neither project. Comparing three extensions, however, will allow us to illustrate how our coverage measure can be applied.

Our starting point is to compile LLM labels for all 220 scenes belonging to $U$, the union of the TRPS, Zhang, LJSP and \acronym data sets. We collected labels for all 23 languages shown in Figure~\ref{fig:llm_evaluation}. For each language $L$, the similarity $\text{sim}_L(i,j)$ between scenes $i$ and $j$ is defined as 1 if the labels for these scenes match and 0 otherwise.
The overall similarity $\text{sim}(i,j)$ between scenes $i$ and $j$ is then defined as the average of $\text{sim}_L(i,j)$ across the 23 languages.

We combined this similarity measure with Equation 1 to compute coverage scores for the TRPS along with three extensions of the TRPS. Table~\ref{tab:coverage_scores} shows that the Zhang and LJSP extensions improve the coverage of the TRPS by a relatively small margin, but the \acronym set achieves a more substantial improvement. All scores in Table~\ref{tab:coverage_scores} are relatively high because the 220 scenes in $U$ are mainly examples of \concept{on} and \concept{in}, and all four sets include examples of these relations. We computed confidence intervals on the coverage scores by creating 1000 bootstrap samples of the universe $U$, and recording the middle 95\% of scores across these samples. The confidence interval for TRPS + \acronym does not overlap any of the others, suggesting that our design goal was achieved, and that this data set provides better coverage of the space of possible relations than the other two extensions of the TRPS.



To visualize the space of semantic relations, we applied multidimensional scaling to the $220 \times 220$ dissimilarity matrix $D$, where $D_{ij} = 1 - \text{sim}(i,j)$. The resulting two-dimensional representation of the 220 scenes is shown in Figure~\ref{fig:scene_space}. The four different panels show how different sets of stimuli are distributed across the space, and are consistent with our quantitative finding that the \acronym set provides greater coverage than either the Zhang set or the LJSP set. The first MDS dimension can be interpreted as a cline from \concept{on} to \concept{in}, but the second dimension does not appear to be interpretable, and a stress plot suggests that at least three dimensions are needed to account well for the dissimilarity matrix $D$. It therefore makes more sense to compute coverage scores using the original dissimilarity $D$ (as we did for Table~\ref{tab:coverage_scores})
than using distances based on the MDS solution in Figure~\ref{fig:scene_space}.

The core idea behind our coverage measure can also be used to quantify the extent to which individual scenes go beyond the TRPS. We ranked the 42 images in the \acronym set based on the similarity of each image to its closest neighbour in the TRPS. The results suggest that scenes illustrating ``under'' (including table under cup and head under hat) are the scenes most similar to the TRPS, and in particular to the TRPS scene with ``spoon under cloth.'' Nine of the new scenes have zero similarity to every TRPS scene, and therefore depart maximally from the TRPS. These scenes include scenes illustrating the topological relation ``outside'' (including fish outside fishbowl, and bird outside birdcage) and scenes showing cardinal relations (including ``New Zealand east of Australia''). These rankings suggest which individual scenes are most useful to include in future behavioral experiments.

The \acronym set, however, is relatively small, which means that it can be easily labeled in its entirety. We collected labels from 10 native Chinese speakers and 7 native English speakers, and have released them at the OSF along with images of the stimuli. The scenes were often but not always successful at eliciting the target relations, and the modal response aligned with the target for 37 (Chinese) and 33 (English) out of 42 scenes. Importantly, the modal human response did not appear among the TRPS labels collected by \citet{carstensenkhr19} for 13 (Chinese) and 15 (English) out of 42 scenes. This finding confirms that the \acronym set does succeed in extending the coverage of the TRPS.

\section{Adding languages to existing data sets}

\begin{figure}[t]
\begin{center}
\includegraphics[scale=0.9]{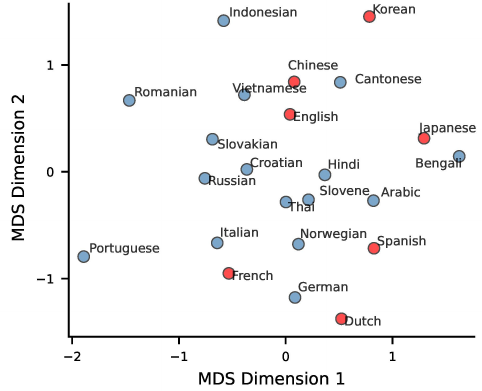}
\end{center}
\vspace{-0.2in}
\caption{MDS visualization of the similarity between spatial systems from the languages represented in Figure \protect\ref{fig:llm_evaluation}. Languages appearing in the \protect\citet{carstensenkhr19} data set are shown in red.}
\vspace{-0.1in}
\label{fig:language_space}
\end{figure}

Suppose we would like to decide which of the languages in Figure~\ref{fig:llm_evaluation}b increases the coverage of the seven-language \citet{carstensenkhr19} data by the greatest extent. The similarity measure required by Equation 1 can be defined as $1-D$, where $D$ is a 23 by 23 matrix of distances between all of the languages in Figure~\ref{fig:llm_evaluation}. We define the distance between any pair of languages using variation of information~\citep{meila07}, an information-theoretic metric that quantifies the difference between the partitions induced by LLM labels for these two languages. Figure~\ref{fig:language_space} shows a two-dimensional MDS visualization of distances between these languages. The distance matrix suggests, for example, that Cantonese may be a  relatively low priority because its system is relatively similar to the Chinese system already documented in the Carstensen et al.\ data. If we order languages based on their distance from their nearest neighbour in the Carstensen et al.\ set, Portuguese and Romanian emerge as the languages most distant from any of the seven Carstensen et al.\ languages, and the two that increase coverage by the greatest extent.

To test this prediction we computed distance matrices $D$ separately based on LLM labels and the human labels for the 21 languages in the \citet{xuk10} data set. Five of the Carstensen et al.\ languages are included in these matrices, and for each of the remaining 16 languages we extracted its distance to the nearest Carstensen et al.\ language. Distances based on LLM and human data had a Pearson correlation of 0.49 (95\% CI [0.19, 0.83]), where the confidence interval is based on 1000 bootstrapped samples of the 21 language set. Portuguese and Romanian (the two targets identified using the LLM data) appear among the four languages most distant from the Carstensen et al.\ languages according to the Xu and Kemp data. These results provide tentative support for the idea that LLM data can help identify languages that increase the coverage of existing data sets.



\section{Discussion}

We showed that LLMs align relatively well with humans when asked to label spatial relations in high-resource languages. This alignment suggests that LLMs can contribute to research on spatial categorization, and we illustrated how LLM labels can be used to choose scenes and languages that extend the coverage of existing spatial categorization data sets. Our approach can potentially be used to scale up to a data set including the roughly 80 languages represented on \texttt{prolific.com} and hundreds of scenes. For this purpose, the set of target languages is already clear, but choosing which scenes to include remains a difficult challenge. One possibility is to generate thousands of scenes using the feature-based approach of \citet{carstensenxkr15}, then apply our coverage score to LLM labels for these scenes to identify a smaller subset to include in the actual experiment.

Instead of working with features, we filled gaps in the TRPS by identifying spatial terms in English and Chinese that lacked representation. This approach can easily be extended to include missing terms from other languages, but does not guarantee systematic coverage of the space of possible relations. In contrast, a feature-based approach can enumerate all logically possible configurations of features and then select a subset that provides both broad and even coverage of the entire space. Combining our LLM-based approach with a feature-based approach is therefore a high priority for future work.

Although we focused on applying our coverage measure to LLM labels, this measure can usefully be applied to human labels. One possible application might consider the data set of \citet{levinsonm03}, which includes TRPS labels in nine typologically diverse languages, four of which are not supported by Google Translate. Our coverage measure could be used to compare the coverage of this set relative to a larger set that includes only high-resource languages. This comparison would provide some information about how much diversity is missed by LLM-based approaches that work only with high-resource languages.

Our approach belongs to the tradition of prior experimental work using the TRPS, but another line of work uses textual corpora to study cross-linguistic variation in spatial categorization using textual corpora~\citep{viechnicki2025spatial,beekhuizen2025spatial}. These two approaches are complementary and LLMs can contribute to both: for example, LLMs may be useful for extracting spatial relation markers from multilingual corpora. 
As LLMs continue to improve, we expect that they will prove increasingly useful for both experimental and corpus-based work on spatial categorization across languages.

\section{Acknowledgments}
We thank \citet{zhang13} for sharing her stimuli.  
Generative AI was used as described in the paper to label images and to generate images illustrating spatial relations. The \acronym data set is available at \url{https://doi.org/10.5281/zenodo.18765858}

\balance
\printbibliography

\end{document}